\documentclass{bmvc2k}

\usepackage{amsmath}
\usepackage{amssymb}
\usepackage{multirow}
\usepackage{booktabs}
\usepackage{xcolor}
\usepackage{caption}
\DeclareCaptionFont{myblue}{\color[rgb]{0,0,0.5}}
\title{Seg-HGNN: Unsupervised and Light-Weight Image Segmentation with Hyperbolic Graph Neural Networks}

\addauthor{Debjyoti Mondal}{d.mondal@samsung.com}{1}
\addauthor{Rahul Mishra}{mishra.rl@samsung.com}{1}
\addauthor{Chandan Pandey}{chandan.p@samsung.com}{1}

\addinstitution{
Samsung R\&D Institute India\\
Bangalore
}

\runninghead{Mondal, Mishra, Pandey}{Seg-HGNN}

\begin{document}

\maketitle

% 0_abstract.tex
\begin{abstract}
    Image analysis in the euclidean space through linear 
    hyperspaces is well studied. However, in the quest for 
    more effective image representations, we 
    turn to hyperbolic manifolds. They 
    provide a compelling alternative to capture 
    complex hierarchical relationships in images with 
    remarkably small dimensionality. 
    To demonstrate hyperbolic embeddings' competence,
    we introduce a light-weight hyperbolic graph neural network for image segmentation,
    encompassing patch-level features in a very small embedding size.
    Our solution, Seg-HGNN, surpasses the current best unsupervised
    method by 2.5\%, 4\% on VOC-07, VOC-12 for localization, and by 0.8\%, 1.3\% on
    CUB-200, ECSSD for segmentation, respectively. 
    With less than 7.5k trainable parameters, Seg-HGNN delivers effective and fast ($\approx 2$ images/second) results on very standard GPUs like the GTX1650. 
    This empirical evaluation presents 
    compelling evidence of the efficacy and potential of 
    hyperbolic representations for vision tasks.
\end{abstract}

% 1_intro.tex
\section{Introduction}
\label{sec:intro}
Image segmentation and object localization are crucial
tasks with diverse applications. Accurately tracing objects
in images and pinpointing their spatial coordinates are essential
in fields such as robotics, medical imaging, and augmented reality.

Traditional methods, primarily in the Euclidean space, have made 
notable progress in these tasks \cite{chen2021transunet, long2015fully, ronneberger2015u}.
However, with the increasing complexity and volume of visual data,
novel approaches are needed for efficiency, scalability, and richer insights.

Images naturally have latent local hierarchies.
Hyperbolic geometry, with its capacity to capture hierarchical and 
tree-like structures, is particularly well-suited for complex, 
interconnected visual data \cite{doi:10.1142/5914, khrulkov2020hyperbolic}. 
Although hyperbolic operations are expensive, recent advancements \cite{dai2021hyperbolic}
have reduced compute needs to a large extent.

In the era of data-driven techniques and large-scale image analysis, 
our method prioritizes reducing computational and memory demands while 
being unsupervised and in the hyperbolic realm. With minimal dimensionality and a 
test-time training approach, our solution is ideal for real-time 
applications on resource-constrained edge devices. Our work paves 
the way for more efficient and accessible image analysis solutions, 
extending beyond segmentation and object localization.

Through this work, we make the following contributions:
\begin{enumerate}
    \item \textbf{Hyperbolic GNNs for Image Analysis:} 
    We present Seg-HGNN, a novel framework for image segmentation 
    that captures latent structures within images, while being constrained 
    on a hyperbolic manifold. Leveraging Hyperbolic Graph 
    Neural Networks (HGNNs), we show how this non-Euclidean framework 
    improves segmentation and object localization.
    \item \textbf{Richer image representations in low dimensions:}
    We show a method of achieving locally aware and 
    semantically rich image representations in low embedding sizes and benchmark their preformance. 
\end{enumerate}

% 2_formatting.tex
\section{Related Work}
\label{sec:related}

\paragraph{Image segmentation.} It is the process of dividing an image into 
multiple distinct regions or segments, each of which 
corresponds to a meaningful object. Deep Convolutional 
Neural Networks (CNNs) \citep{long2015fully, sermanet2013overfeat} 
have shown promising performance for this task and
are the go-to backbone of recent methods. 
The coming of encoder-decoder models
\citep{badrinarayanan2017segnet, ronneberger2015u, chen2017deeplab} 
have given way to numerous state-of-the-art variants for segmentation tasks.

While transformers \citep{vaswani2017attention} revolutionized language processing,
Vision Transformers (ViT) 
\citep{dosovitskiy2020image} were shown to be at-par with CNNs
\citep{strudel2021segmenter, hatamizadeh2022unetr}.
However, these models need a huge amount of training data.
\citet{caron2021emerging} trained ViTs using self-DIstillation 
with NO labels (DINO), and it was seen that the generated attention 
maps corresponded to semantic segments in the image. 
Works of \citet{melas2022deep, shi2000normalized}
explore classical graph theory, using deep features for localization
and segmentation. 
Seg-HGNN takes these ideas into the hyperbolic manifold, while being lightweight
and unsupervised.

\paragraph{Graph Convolutional Networks. (GCNs)} GCNs \citep{kipf2017semisupervised} are 
networks that take advantage of structured data, and 
have achieved remarkable results in complex tasks like drug discovery 
\citep{you2018graph} and protein analysis \citep{jumper2021highly}.
DGCNet \citep{zhang2019dual} utilizes a dual GCN framework to model 
image feature context in both coordinate and feature spaces, and merges them
back together. \citet{hu2020class}
introduces a class-wise dynamic GCN module to cluster similar pixels
together and dynamically aggregate features. 
DeepCut \citep{aflalo2023deepcut} constructs a graph with the pair-wise affinities 
between local image features and
performs correlation clustering. Seg-HGNN also tries to use this expressivenes of GCNs.

\paragraph{Hyperbolic spaces.} 
Known for embedding hierarchies and tree-like structures
with minimal distortion in low dimensions \citep{nickel2017poincare}, they have
inspired hyperbolic variants of neural network blocks 
\citep{ganea2018hyperbolic, shimizu2020hyperbolic}.
\citet{khrulkov2020hyperbolic} shows the presence of hierarchies in
image datasets, which brought along early
success in hyperbolic computer vision for few-shot, zero-shot, and 
unsupervised learning \citep{fang2021kernel, park2021unsupervised, yan2021unsupervised}.
Exploiting the intrinsic negative curvature of hyperbolic manifolds,
they emerge as a powerful framework for geometric deep learning.
However, many rely on tangent spaces for aggregation and message
passing \citep{liu2019hyperbolic, chami2019hyperbolic}. 
\citet{dai2021hyperbolic} presents a fully hyperbolic GNN 
by introducing a Lorentz linear transform.

\paragraph{Image segmentation in hyperbolic spaces.}
\citet{atigh2022hyperbolic} formulates
dataset-level hierarchy and does pixel-level classification in the hyperbolic space
in a supervised-setup. They highlight the richness of embeddings,
offering boundary information and uncertainty measures in low dimensions.

% 3_finalcopy.tex
\section{Preliminaries}
\label{sec:prelims}
\subsection{Hyperbolic Space}
A Riemannian manifold $(\mathcal{M}, g)$ is a smooth, connected space 
where each point $\boldsymbol{x} \in \mathcal{M}$, has a tangent 
space $\mathcal{T}_{\boldsymbol{x}}\mathcal{M}$, that behaves like it is Euclidean. 
Hyperbolic spaces are a specific type of Riemannian manifold that 
exhibit constant negative curvature \cite{HypBook}. They are studied 
using a few isometric models\cite{FoG}.
We 
choose the Lorentz model for its numerical stability \citep{nickel2018learning}.

The Lorentz model for an $n$-dimensional hyperbolic space is defined by the manifold
$\mathcal{L} = \{\boldsymbol{x} = [x_0, x_1, ..., x_n] \in \mathbb{R}^{n+1} : \langle \boldsymbol{x}, \boldsymbol{x}\rangle_{\mathcal{L}} = -1, x_0 > 0\}$,
with the metric tensor $g = \text{diag}([-1, \mathbf{1}_n^\mathsf{T}])$.
We describe a few needed operations below.

\paragraph{Inner product.} The Lorentz inner product is defined as 
\begin{equation}
    \label{eq:inner_prod}
    \langle \boldsymbol{x}, \boldsymbol{y} \rangle_\mathcal{L} = \boldsymbol{x}^\mathsf{T}g\boldsymbol{y} = -x_0y_0 + \sum_{i = 1}^{n}x_iy_i
\end{equation}

% \noindent{\textbf{Distance.}} Geodesics are defined as the curves with the shortest distance between points, whose lengths are measured in terms of the Riemannian metric. For any $\boldsymbol{x}, \boldsymbol{y} \in \mathcal{L}$, the distance between them is given by
% \begin{equation}
%     \label{eq:distance}
%     d_\mathcal{L}(\boldsymbol{x}, \boldsymbol{y}) = \text{arccosh}(-\langle\boldsymbol{x},\boldsymbol{y}\rangle_\mathcal{L})
% \end{equation}
% where $\text{arccosh}(\cdot)$ is the inverse hyperbolic cosine function.

\paragraph{Exponential and logarithmic maps.} An exponential map $\text{exp}_{\boldsymbol{x}}(\boldsymbol{v})$,
projects a vector $\boldsymbol{v} \in \mathcal{T}_{\boldsymbol{x}}\mathcal{M}$, onto the manifold $\mathcal{M}$.
A logarithmic map, $\text{log}_{\boldsymbol{x}}(\boldsymbol{y})$, is the inverse operation,
which maps a point $\boldsymbol{y} \in \mathcal{L}$ to the tangent space of $\boldsymbol{x}$,
that is $\mathcal{T}_{\boldsymbol{x}}\mathcal{M}$. These functions complement each other 
by satisfying $\text{log}_{\boldsymbol{x}}(\text{exp}_{\boldsymbol{x}}(\boldsymbol{v})) = \boldsymbol{v}.$
For $\boldsymbol{x}, \boldsymbol{y} \in \mathcal{L}$, and
$\boldsymbol{v} \in \mathcal{T}_{\boldsymbol{x}}\mathcal{L}$, they are defined as
\begin{equation}
    \label{eq:exp}
    \text{exp}_{\boldsymbol{x}}(\boldsymbol{v}) = \text{cosh}(|| \boldsymbol{v}||_\mathcal{L})\boldsymbol{x} + \text{sinh}(|| \boldsymbol{v}||_\mathcal{L})\frac{\boldsymbol{v}}{||\boldsymbol{v}||_\mathcal{L}}
\end{equation}
\begin{equation}
    \label{eq:log}
    \text{log}_{\boldsymbol{x}}(\boldsymbol{v}) = \frac{\text{arccosh}(-\langle\boldsymbol{x},\boldsymbol{v}\rangle_\mathcal{L})}{\sqrt{\langle\boldsymbol{x},\boldsymbol{v}\rangle_\mathcal{L}^2 - 1}}(\boldsymbol{v} + \langle\boldsymbol{x},\boldsymbol{v}\rangle_\mathcal{L}\boldsymbol{x})
\end{equation}
where $||\boldsymbol{v}||_\mathcal{L} = \sqrt{\langle\boldsymbol{v},\boldsymbol{v}\rangle_\mathcal{L}}$.

\paragraph{Isometric bijections.} The Poincaré ball $\mathcal{B}$ and the Klein
model $\mathcal{K}$ are two other models of hyperbolic spaces.
There are bijective relations connecting all these model.
For a point $\boldsymbol{x} = [x_0, x_1, ..., x_n] \in \mathcal{L}$,
its corresponding point $\boldsymbol{b} = [b_0, b_1, ..., b_{n-1}] \in \mathcal{B}$,
is obtained as 
\begin{equation}
    \label{eq:l_pb}
    p_{\mathcal{L} \rightarrow \mathcal{B}}(\boldsymbol{x}) = \frac{[x_1, ..., x_n]}{x_0 + 1}, p_{\mathcal{B} \rightarrow \mathcal{L}}(\boldsymbol{b}) = \frac{[1 + ||\boldsymbol{b}||^2, 2\boldsymbol{b}]}{1 - ||\boldsymbol{b}||^2}
\end{equation}
Similarly for $\boldsymbol{x} = [x_0, x_1, ..., x_n] \in \mathcal{L}$ and $\boldsymbol{k} = [k_0, k_1, ..., k_{n-1}]\in \mathcal{K}$, we have
\begin{equation}
    \label{eq:l_k}
    p_{\mathcal{L} \rightarrow \mathcal{K}}(\boldsymbol{x}) = \frac{[x_1, ..., x_n]}{x_0}, p_{\mathcal{K} \rightarrow \mathcal{L}}(\boldsymbol{k}) = \frac{[1, \boldsymbol{k}]}{\sqrt{1 - ||\boldsymbol{k}||^2}}
\end{equation}

\subsection{Hyperbolic Graph Convolutional Networks (GCNs)}
GCNs \cite{kipf2017semisupervised} introduce the convolution operation on structured, graph data.
A graph $\mathcal{G}$ is made of a set of vertices $\mathcal{V}$ and edges $\mathcal{E}$.
Each node $i$ has an associated embedding $\boldsymbol{h}_i^0$ and a set of neighbours, $\mathcal{N}_i$.
The convolution operation for a single layer is done with three underlying sub-operations,
namely, feature transformation,
message-passing, and update. 
Constructing a GCN in the hyperbolic space presents unique challenges due to 
the need to uphold the constraints of hyperbolicity across all layers.
Several strategies have addressed these challenges in notable
works \citep{liu2019hyperbolic, chami2019hyperbolic}. 
They use
eq.(\ref{eq:log}) and eq.(\ref{eq:exp}) to perform transformation and
message passing in the tangent space and then project it back to the manifold.

\paragraph{Hyperbolic Feature Transformation.}
Using the feature transform defined as a matrix-vector multiplication
in the Euclidean space, as in \citet{kipf2017semisupervised}, breaks the hyperbolic
constraint when applied to hyperbolic node representations. 
To avoid this, and to make the aggregation and message-passing steps
take less compute, we adopt the Lorentz linear transform introduced in \citet{dai2021hyperbolic},
which is defined as 
\begin{equation}
    \label{eq:stiefel}
    \begin{split}
    \boldsymbol{y} &= \boldsymbol{W}\boldsymbol{x} \\
    \text{s.t. } \boldsymbol{W} &= \begin{bmatrix} 1 & \boldsymbol{0}^\mathsf{T} \\ \boldsymbol{0} & \boldsymbol{\widetilde{W}} \\ \end{bmatrix}, \boldsymbol{\widetilde{W}}^\mathsf{T}\boldsymbol{\widetilde{W}} = \boldsymbol{I}
    \end{split}
\end{equation}
where they show that $\boldsymbol{x}, \boldsymbol{y} \in \mathcal{L}$ and
$\boldsymbol{\widetilde{W}}$ is constrained to be
on the Stiefel manifold \citep{gao2021riemannian}.

\paragraph{Hyperbolic Neighborhood Aggregation.}
We resort to the Einstein midpoint \citep{doi:10.1142/5914} for aggregating the node representations.
However, since the Einstein midpoint is defined in the Klein model, we use the bijections defined in
eq.(\ref{eq:l_k}).

\citet{dai2021hyperbolic} also tells that applying a
non-linear activation on the Poincaré ball model does not break hyperbolicity,
that is, $\forall \boldsymbol{b} \in \mathcal{B}$, we get $\sigma(\boldsymbol{b}) \in \mathcal{B}$.

In summary, a graph layer $l$ in the hyperbolic space would look like 
\begin{equation}
    \begin{split}
        \boldsymbol{\tilde{k}}_i^l &= p_{\mathcal{L} \rightarrow \mathcal{K}}(\boldsymbol{W}^l\boldsymbol{h}_i^{l-1}) \\
        \boldsymbol{m}_i^l &= p_{\mathcal{K} \rightarrow \mathcal{L}}(\sum_{j \in \mathcal{\tilde{N}}_i}w_{ij}\gamma_j\boldsymbol{\tilde{k}}_j^l / \sum_{j \in \mathcal{\tilde{N}}_i}\gamma_j) \\
        \boldsymbol{h}_i^l &= p_{\mathcal{B} \rightarrow \mathcal{L}}(\sigma(p_{\mathcal{L} \rightarrow \mathcal{B}}(\boldsymbol{m}_i^l)))
    \end{split}
\end{equation}
where $\boldsymbol{h}_i \in \mathcal{L}$, $\boldsymbol{W}$ is the Lorentz linear transformation matrix,
$w_{ij}$ is the weight of the edge between nodes $i$ and $j$, and
$\mathcal{\tilde{N}}_i = \mathcal{N}_i \cup \{i\}$ is the neighborhood of node $i$. 
$\gamma_i = \frac{1}{\sqrt{1 - ||\boldsymbol{k}_i||^2}}$ is called the Lorentz factor, and is used during the calculation of the Einstein midpoint.

\subsection{Segmentation as a Graph Clustering problem}
The image is divided into patches and each one is a vertex in graph
$\mathcal{G} = (\mathcal{V}, \mathcal{E})$. 
We model the task of segmentation as partitioning 
these patches into $k$ disjoint clusters $C_1, C_2, ..., C_k$ such that $\cup_iC_i = \mathcal{V}$.
We use the Normalized Cuts \citep{shi2000normalized} objective, and for a partition $\mathcal{P}$ of 
graph $\mathcal{G}$, the Normalized-Cut cost is given by 
\begin{equation}
    \label{eq:ncut}
    \text{Ncut}(\mathcal{P}) = \frac{\text{cut}(\mathcal{P})}{\text{assoc}(\mathcal{P})} + \frac{\text{cut}(\mathcal{\bar{P}})}{\text{assoc}(\mathcal{\bar{P}})}
\end{equation}
where $\text{cut}(\cdot)$ is the weight of all removed edges 
between the partitions, $\text{assoc}(\cdot)$ is the total 
weight of partition $\mathcal{P}$, and $\bar{\mathcal{P}}$ is the other partition.
In practice, the algorithm recursively bisects the graph using the eigen
vectors of the graph Laplacian matrix to form a binary tree structure, 
representing a hierarchical clustering of the graph. The partitioning process 
continues until a stopping criterion is met.

% \begin{figure}
%     \centering
%     \includegraphics[width = \linewidth]{einstein.pdf}
%     \caption{Einstein midpoint on the Klein model with points $a, b$ and $c$. 
%     Here, $\gamma_i$ is the Lorentz factor for node $i$.}
% \end{figure}

% 4_method.tex
\section{Method}
\label{sec:method}

\begin{figure*}
    \centering
    \includegraphics[width = \linewidth]{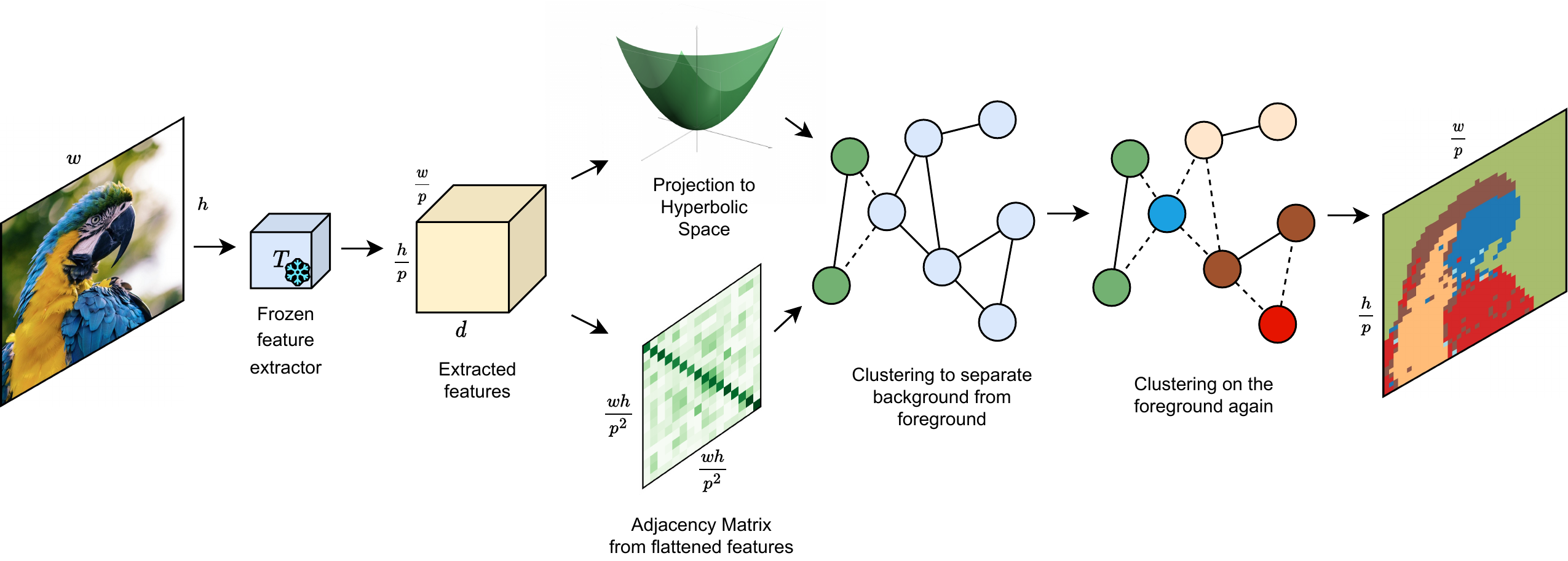}
    \caption{\textbf{Method overview.} We extract patch-level features from a
    frozen model $\boldsymbol{T}$, flattening them to $\boldsymbol{f}$. 
    Edge weights are obtained from the Gram matrix(eqn.(\ref{eq:normalization})) of $\boldsymbol{f}$.
    We project $\boldsymbol{f}$ to the lorentz manifold
    to obtain initial node embeddings. We optimize an unsupervised graph partitioning loss
    to obtain $k$ clusters.}
\end{figure*}

% We use a pre-trained network to get the initial node representations,
% form the graph in the hyperbolic space, and then use a clustering loss to
% train the graph. The same framework is used for segmentation and localization.
% We outline the key steps in this section.

\subsection{Patch-level features}
We use a vision transformer network \citep{dosovitskiy2020image} $T$,
to get the patch-level features. For an image with size $m \times n$, we 
get $\frac{mn}{p^2}$ patches, where $p$ is the patch size of transformer $T$.
\citet{amir2021deep}, and \citet{melas2022deep} have shown that
the $key$ matrix from the last layer of $T$ exhibits superior 
performance in various tasks. Therefore, the image features  
are represented as $\boldsymbol{f} \in \mathbb{R}^{\frac{mn}{p^2} \times d}$, 
containing the $key$-values of embedding size $d$ for each patch.

\subsection{Obtaining Hyperbolic features}
The obtained features $\boldsymbol{f}$ are in the Euclidean
space and need to be exported to the lorentz space $\mathcal{L}$. 
% Let $\boldsymbol{f}_i$ be one of these feature vectors, 
Let $\boldsymbol{o}_\mathcal{L} := [1, 0, ..., 0]$ be called the origin in $\mathcal{L}$.
From the definition in eq.(\ref{eq:inner_prod}), we see that 
$\langle \boldsymbol{o}_\mathcal{L}, [0, \boldsymbol{f}_i] \rangle_\mathcal{L} = 0$.
This allows us to think of $[0, \boldsymbol{f}_i]$ as a vector in the tangent space of $\boldsymbol{o}_\mathcal{L}$. We can then just use the exponential map defined in eq.(\ref{eq:exp}) to get representations on $\mathcal{L}$.

\subsection{Edge weights and Clustering loss}
\label{subsec:loss}
The edge weights are obtained from the correlation matrix of the 
transformer features, $\boldsymbol{f}\boldsymbol{f}^\mathsf{T}$. An 
additional normalizing factor comes from the idea of normalized graph 
Laplacian \citep{belkin2006manifold, zhu2003semi}. For an 
edge and its weight $\tilde{w}_{ij}$, we divide it by the square root 
of the degrees of the associated nodes. Also, as normalized-cut works with 
positive weights only, we threshold these values at 0.
\begin{equation}
    \label{eq:normalization}
    w_{ij} = \frac{\max(0, \tilde{w}_{ij})}{\sqrt{|\tilde{\mathcal{N}}_i||\tilde{\mathcal{N}}_j|}} 
    = \frac{\max(0, \tilde{w}_{ij})}{\frac{mn}{p^2}}
\end{equation}
After passing through a one-layer hyperbolic GCN, we use a 
fully-connected layer followed by a $\text{softmax}(\cdot)$, to get the probabilities of cluster assignment for each patch, as a matrix $\mathcal{S}$. 
\begin{equation}
    \label{eq:softmax}
    \begin{split}
    \mathcal{F} &= \text{HyperbolicGCN}({\boldsymbol{f}^\mathcal{L}}, \boldsymbol{w}) \\
    \mathcal{F}^\prime &= \text{log}_{\boldsymbol{o}_\mathcal{L}}(\mathcal{F}) \\ 
    \mathcal{F}^{\prime\prime} &= \text{FullyConnected}(\mathcal{F}^\prime) \\
    \mathcal{S} &= \text{softmax}(\mathcal{F}^{\prime\prime})
    \end{split}
\end{equation}
We use the relaxed normalization-cut proposed in \citet{bianchi2020spectral}, to get the clustering loss.
\begin{equation}
    \label{eq:loss}
    loss_{\text{N-cut}} = -\frac{\text{tr}(\mathcal{S}^\mathsf{T}\boldsymbol{w}\mathcal{S})}{\text{tr}(\mathcal{S}^\mathsf{T}\boldsymbol{D}\mathcal{S})} + \left \Vert \frac{\mathcal{S}^\mathsf{T}\mathcal{S}}{||\mathcal{S}^\mathsf{T}\mathcal{S}||_F} - \frac{\boldsymbol{I}_k}{\sqrt{k}}\right\Vert_F
\end{equation}
where $\text{tr}(\cdot)$ is the trace function, $\boldsymbol{D} = \text{diag}(\sum_j w_{ij})$, $k$ is the number of clusters we want to segment into, and $\boldsymbol{I}$ is the identity matrix.

\subsection{Optimization on Hyperbolic manifolds}
We keep the transformer $T$ frozen. 
$\text{HyperbolicGCN}(\cdot)$ and the $\text{FullyConnected}(\cdot)$ layers contain the only
trainable parameters. The operations and parameters in the latter layer are euclidean,
and can be learned by any standard gradient descent optimizer.
The transformation matrix $\boldsymbol{W}$ in eq.(\ref{eq:stiefel}) is bounded by
the orthogonality contraint of submatrix $\boldsymbol{\widetilde{W}}$.
The set of matrices with orthonormal columns form another Riemannian 
manifold called the Stiefel manifold \citep{boothby1986introduction}. 
We use Riemannian stochastic
gradient descent optimizer to learn $\boldsymbol{\widetilde{W}}$.

\subsection{Segmentation and Localization}
\label{subsec:tasks}
The cluster assignment probabilites ($\mathcal{S}$) and
the cluster count ($k$) facilitate segmentation and localization.
For object localization, we set $k$ as 2 and color-map the patches to assigned clusters.
We draw a bounding box around clusters with an area greater than 
that of 4 patches. Following \citet{aflalo2023deepcut}, the cluster which
apprears on more than 2 edges is called the background.
The method used for Object Segmentation is identical to Localization, except the
inclusion of bounding boxes.
For Semantic Part Segmentation, we use a top-down recursive approach.
After seperating the foreground patches
from the image with $k = 2$, we perform another round of clustering 
with $k = 4$ on the foreground patches only. This also
reduces the bias of clustering towards large clusters.

% 5_experiments.tex
\section{Experiments and Results}
\label{sec:exp}

\subsection{Training Details}
We use DINO \cite{caron2021emerging} trained
ViT-S with a patch size of 8 as the feature extractor $T$.
These features are projected to the 16-dimensional 
Lorentz space, unless stated otherwise.
The proposed $\text{HyperbolicGCN}(\cdot)$ has one graph layer,
with both input and output dimension size of 16.
The $\text{FullyConnected}(\cdot)$ layer follows with a stack of two
linear layers with hidden-layer dimension 32 and output
size of $K = 2$ or $4$, for tasks in Section \ref{subsec:tasks}.

We use a test-time training paradigm, where patches from 
the image to be segmented are clustered by the hyperbolic graph framework.
Our model has atmost 7.3k parameters on the euclidean,
and 256 parameters on the stiefel manifold. This totals to 
\textbf{7.5k trainable parameters}, compared to 30k parameters in the
current state-of-the-art \cite{aflalo2023deepcut}.

With the relaxed normalization-cut loss, we train our model for 10 epochs for localization and 
object segmentation, and for 100 epochs for semantic
part segmentation. We use a learning rate of 0.01 for the euclidean
parameters and 0.1 for the stiefel parameters.
More implementation details and perfomance analysis are provided in the 
supplementary material.

\subsection{Results}
In this sub-section, we assess Seg-HGNN's performance on popular 
benchmarks for localization and segmentation. We compare it against 
unsupervised and some supervised methods.

\begin{table}[]
    \centering
    \fontsize{9}{10}\selectfont
    \begin{tabular}{@{}l|cc|ccc@{}}
        \toprule
        \multirow{2}{*}{\textbf{Method}}              & \multicolumn{2}{c}{\textbf{Object Localization}} & \multicolumn{3}{|c}{\textbf{Object Segmentation}}                                                     \\
                                                      & \textbf{VOC-07}                                  & \textbf{VOC-12}                                   & \textbf{CUB-200} & \textbf{DUTS} & \textbf{ECSSD} \\ \midrule
        Selective Search \cite{uijlings2013selective} & 18.8                                             & 20.9                                              & -                & -             & -              \\
        EdgeBoxes \cite{zitnick2014edge}              & 31.1                                             & 31.6                                              & -                & -             & -              \\
        DINO-[CLS] \cite{caron2021emerging}           & 45.8                                             & 46.2                                              & -                & -             & -              \\
        LOST \cite{simeoni2021localizing}             & 61.9                                             & 64.0                                              & -                & -             & -              \\
        OneGAN \citep{benny2020onegan}                & -                                                & -                                                 & 55.5             & -             & -              \\
        \citet{voynov2021object}                      & -                                                & -                                                 & 68.3             & 49.8          & -              \\
        Spectral Methods \cite{melas2022deep}         & 62.7                                             & 66.4                                              & 76.9             & 51.4          & 73.3           \\
        TokenCut \cite{wang2022self}                  & 68.8                                             & 72.1                                              & -                & 57.6          & 71.2           \\
        DeepCut \citep{aflalo2023deepcut}             & 69.8                                             & 72.2                                              & 78.2             & \textbf{59.5} & 74.6           \\ \midrule
        Seg-HGNN (ours)                               & \textbf{72.3}                                    & \textbf{76.1}                                     & \textbf{79.0}    & 57.6          & \textbf{75.9}  \\ \bottomrule
    \end{tabular}
    \caption{Results comparing Object Localization and Object Segmentation performance.}
    \label{tab:results}
\end{table}

\paragraph{Localization and Object Segmentation.}
In Table \ref{tab:results}, we report the object localization performance 
of our unsupervised approach on PASCAL VOC 2007 \cite{pascal-voc-2007} and
2012 \cite{pascal-voc-2012} datasets. Here, we use the 
Correct Localization (CorLoc) metric,
which is the percentage share of images where 
intersection-over-union with the ground truth bounding box 
is greater than 0.5.

Table \ref{tab:results} also compares 
Seg-HGNN's segmentation performance on three datasets :
CUB (Caltech-UCSD Birds-200-2011) \cite{WahCUB_200_2011},
a widely used dataset containing images of birds for 
single object segmentation and semantic part segmentation
with 5794 test images, DUTS \cite{wang2017learning} with 5019 test images, 
and ECSSD \cite{yan2013hierarchical}
with 1000 images.
We report our results using mean Intersection-over-Union (mIOU).

\begin{figure}
    \centering
    \includegraphics[width = 0.6\linewidth]{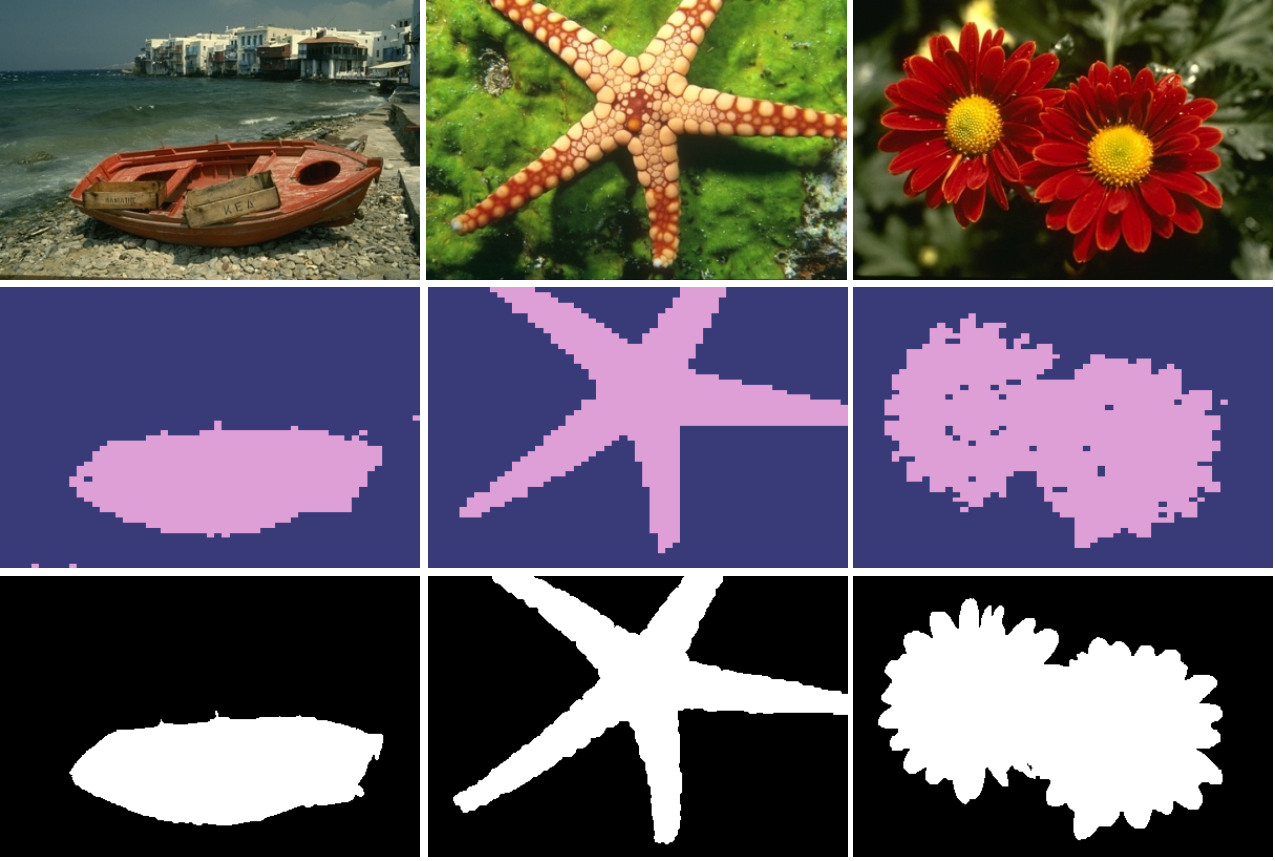}
    \caption{A few samples comparing the quality of segmentation
    achieved by Seg-HGNN. Here, the second row shows the predicted
    masks and the third row has the ground-truth.}
    \label{fig:seg_samples}
\end{figure}

\begin{figure}
    \centering
    \includegraphics[width = 0.6\linewidth]{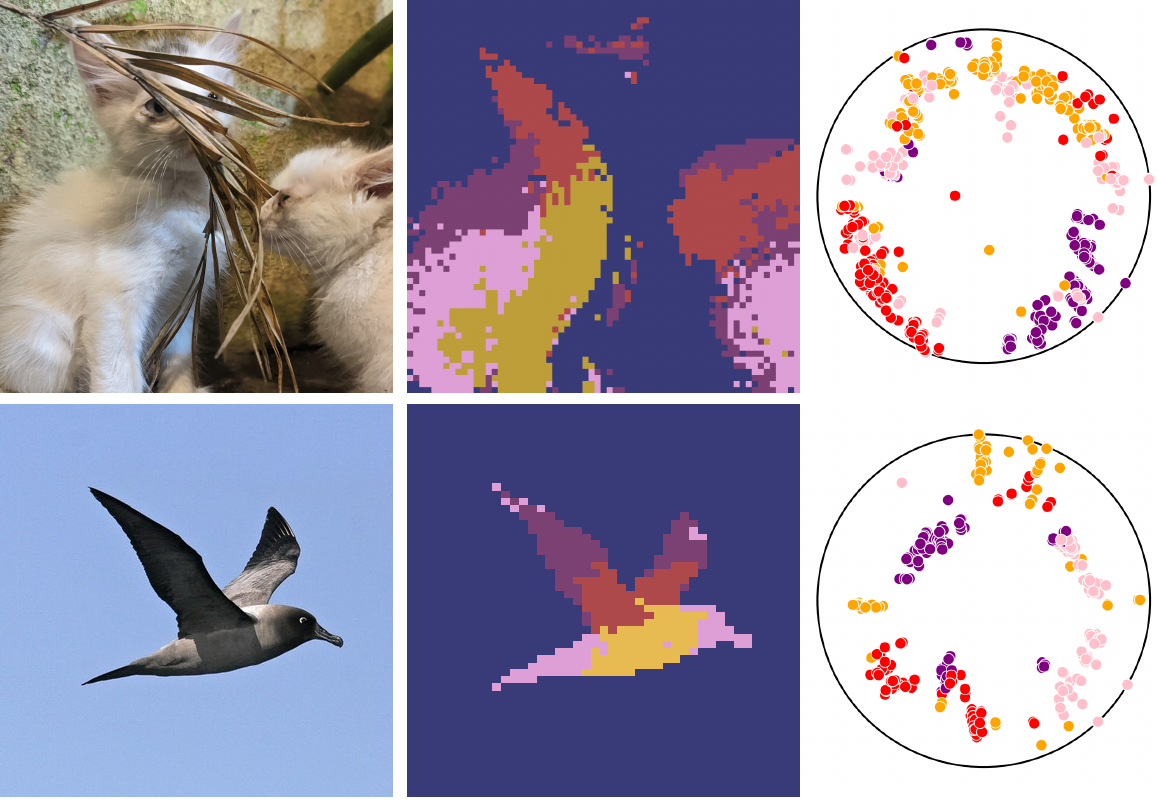}
    \caption{Samples showing the quality of semantic segmentation. 
    The first image shows that even for multiple objects, the
    underlying semantic behind clustering remains intact. 
    For example, the heads of both the cats are clustered together in red.
    The projected hyperbolic embeddings also show the quality of clustering.}
    \label{fig:sem_samples}
\end{figure}

\paragraph{Semantic Part Segmentation.}
\begin{table}[]
    \centering
    \fontsize{9}{10}\selectfont
    \begin{tabular}{@{}lcc@{}}
    \toprule
    \textbf{Method} & \textbf{NMI}         & \textbf{ARI}         \\ \midrule
    SCOPS \cite{hung2019scops} & 24.4                 & 7.1                  \\
    \citet{huang2020interpretable} & 26.1                 & 13.2                 \\
    \citet{choudhury2021unsupervised} & 43.5                 & 19.6                 \\ \midrule
    DFF \cite{collins2018deep} & 25.9                 & 12.4                 \\
    \citet{amir2021deep} & 38.9                 & 16.1                 \\
    DeepCut \cite{aflalo2023deepcut} & \textbf{43.9}                 & 20.2                 \\ \midrule
    Seg-HGNN (ours) & 42.3 & \textbf{20.8} \\ \bottomrule
    \end{tabular}
    \caption{\textbf{Semantic Segmentation results} over the CUB-200 dataset.
    Here, the first three methods use ground truth masks for supervision.}
    \label{tab:semantic}
\end{table}
In Table \ref{tab:semantic} we present the performance of
Seg-HGNN on the CUB dataset for semantic part segmentation.
We use the Normalized Mutual Information (NMI), 
and the Adjusted Rank Index (ARI) scores here.
Seg-HGNN is at par with 
all other unsupervised methods, 
including some supervised methods \cite{hung2019scops, huang2020interpretable, choudhury2021unsupervised}.
Figure \ref{fig:sem_samples}
shows a few samples and corresponding projections of the segmented part embeddings.
These are direct projections from the Poincaré ball model.
We can see that Seg-HGNN tries to put semantically similar parts together.

\subsection{Effectiveness of low-dimensional embeddings}
\begin{figure*}
    \centering
    \includegraphics[width = \linewidth]{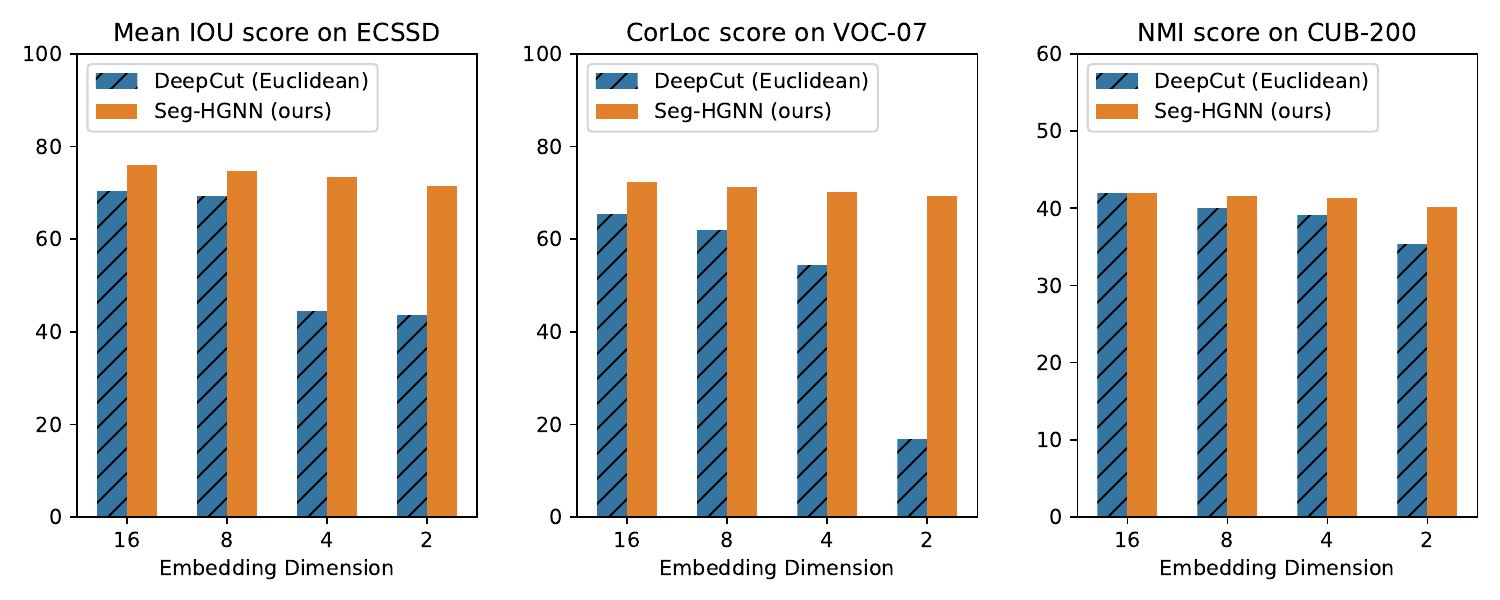}
    \caption{\textbf{Low-dimensional performance of hyperbolic embeddings.}}
    \label{fig:embedding_dimension}
    \vspace{-9pt}
\end{figure*}

Figure \ref{fig:embedding_dimension} presents the effectiveness of
hyperbolic embeddings in a low-dimensional setting.
While performance remains similar at higher dimensions,
there's a drastic drop in Euclidean scores. 
Hyperbolic embeddings can hold information 
for sizes as low as $d = 2$.
The results emphasize that low dimentional hyperbolic embeddings
might be a go to solution when explainability, reduced complexity
and memory footprint are of concern.

\subsection{Comparing used resources}
Hyperbolic operations are said to be compute-demanding. Since even with $d = 2$
we get reliable perfomance, Table \ref{tab:comparison} suggests faster inference
speeds and low resource-usage.
\begin{table}[h!]
    \centering
    \fontsize{8}{9}\selectfont
    \begin{tabular}{@{}l|cccc|cccc@{}}
        \toprule
        \multirow{2}{*}{\textbf{Model}} & \multicolumn{4}{c|}{\textbf{V-RAM (GB) $\downarrow$}} & \multicolumn{4}{c}{\textbf{Infer Rate (img/sec) $\uparrow$}}                                                                                 \\ \cmidrule(l){2-9}
                                        & \textbf{d = 2}                                        & \textbf{4}                                                   & \textbf{8} & \textbf{16} & \textbf{2} & \textbf{4} & \textbf{8} & \textbf{16} \\ \cmidrule(r){1-9}
        \textbf{Seg-HGNN}               & 1.2                                                   & 1.3                                                          & 1.5        & 1.83        & 2.1        & 1.9        & 1.85       & 1.54        \\
        \textbf{DeepCut}                & \multicolumn{4}{c|}{1.81 (d = 64)}                    & \multicolumn{4}{c}{1.44 (d = 64)}                                                                                                            \\ \bottomrule
    \end{tabular}
    \caption{Resource usage and inference rates on 
    ECSSD, including pre and post-processing.}
    \label{tab:comparison}
\end{table}
\vspace{-12pt}
\begin{table}[]
    \centering
    \fontsize{8}{9}\selectfont
    \begin{tabular}{@{}ll|cc@{}}
        \toprule
        \multicolumn{2}{@{}l|}{\textbf{Method}}        & \textbf{VOC-07}  & \textbf{ECSSD}                 \\ \cmidrule(l){1-4}
        \multirow{2}{*}{MoCo-V3 \cite{chenempirical}}  & without Seg-HGNN & 46.19          & 36.7          \\
                                                       & with Seg-HGNN    & \textbf{61.2}  & \textbf{53.3} \\ \cmidrule(l){1-4}
        \multirow{2}{*}{DINO \cite{caron2021emerging}} & without Seg-HGNN & 45.8           & 51.2          \\
                                                       & with Seg-HGNN    & \textbf{72.3}  & \textbf{75.9} \\ \bottomrule
    \end{tabular}
    \caption{The effect of adding Seg-HGNN on top of pretrained methods.}
    \label{tab:pretrain}
\end{table}

\subsection{Effect of adding Seg-HGNN and Conclusion}
We benchmark the performance
of a few unsupervised methods, both with and without Seg-HGNN.
Table \ref{tab:pretrain} shows that adding Seg-HGNN
improves scores for both tasks.
We conclude that Seg-HGNN and hyperbolic manifolds effectively 
represent complex hierarchical structures even in minimal dimensions.

%-------------------------------------------------------------------------

\bibliography{egbib}
\end{document}